\documentclass[11pt,a4paper]{article}
\usepackage[hyperref]{emnlp-ijcnlp-2019}
\usepackage{times}
\usepackage{latexsym}
\usepackage{graphicx}
\usepackage{multicol}
\usepackage{amsmath}
\usepackage{multirow}
\usepackage{booktabs}
\usepackage{subcaption}
\usepackage[ruled, linesnumbered]{algorithm2e}
\usepackage{url}
\usepackage{comment}
\usepackage{siunitx}
\usepackage{amssymb}
\usepackage{bm}

\usepackage{url}

\usepackage{todonotes}   % insert [disable] to disable all notes
\usepackage{tikz}
\usepackage{pgfplots}
\aclfinalcopy % Uncomment this line for the final submission

\title{FewRel 2.0: Towards More Challenging Few-Shot Relation Classification}

\author{Tianyu Gao$^{1}$, Xu Han$^{1}$, Hao Zhu$^{1,2}$, Zhiyuan Liu$^{1}$\thanks{\quad Corresponding author}\hspace{0.5em}, Peng Li$^{3}$, \\\textbf{Maosong Sun$^{1}$, Jie Zhou$^{3}$}\\
$^1$Department of Computer Science and Technology, Tsinghua University, Beijing, China\\
$^2$Carnegie Mellon University, Pittsburgh, PA, USA\\
$^3$Pattern Recognition Center, WeChat AI, Tencent Inc, China\\
{\tt \{gty16,hanxu17\}@mails.tsinghua.edu.cn,zhuhao@cmu.edu}\\
{\tt liuzy@tsinghua.edu.cn,patrickpli@tencent.com}\\
{\tt sms@tsinghua.edu.cn,withtomzhou@tencent.com}\\
}

\date{}

\begin{document}
\maketitle
\begin{abstract}
We present FewRel 2.0, a more challenging task to investigate two aspects of few-shot relation classification models: (1) Can they adapt to a new domain with only a handful of instances? (2) Can they detect none-of-the-above (NOTA) relations? To construct FewRel 2.0, we build upon the FewRel dataset \cite{han2018fewrel} by adding a new test set in a quite different domain, and a NOTA relation choice. With the new dataset and extensive experimental analysis, we found (1) that the state-of-the-art few-shot relation classification models struggle on these two aspects, and (2) that the commonly-used techniques for domain adaptation and NOTA detection still cannot handle the two challenges well. Our research calls for more attention and further efforts to these two real-world issues. All details and resources about the dataset and baselines are released at \url{https://github.com/thunlp/fewrel}.
\end{abstract}

\section{Introduction}

Few-shot learning, which requires models to handle new classification tasks with only a handful of training instances, has drawn much attention in recent years \cite{ravi2016optimization,vinyals2016matching,munkhdalai2017meta,snell2017prototypical}. To advance this field in NLP, \newcite{han2018fewrel} propose FewRel, a large-scale dataset to explore few-shot learning in relation classification. Many efforts~\cite{gao2019hybrid,soares2019matching} have been devoted to the new task and some of the methods even exceed human performance\footnote{\url{https://thunlp.github.io/fewrel.html}} on FewRel. Based on the dataset FewRel, we propose FewRel 2.0, a new task containing two real-world issues that FewRel ignores: (1) few-shot domain adaptation, and (2) few-shot none-of-the-above detection.

\begin{figure}
    \centering
    \includegraphics[width=0.235\textwidth]{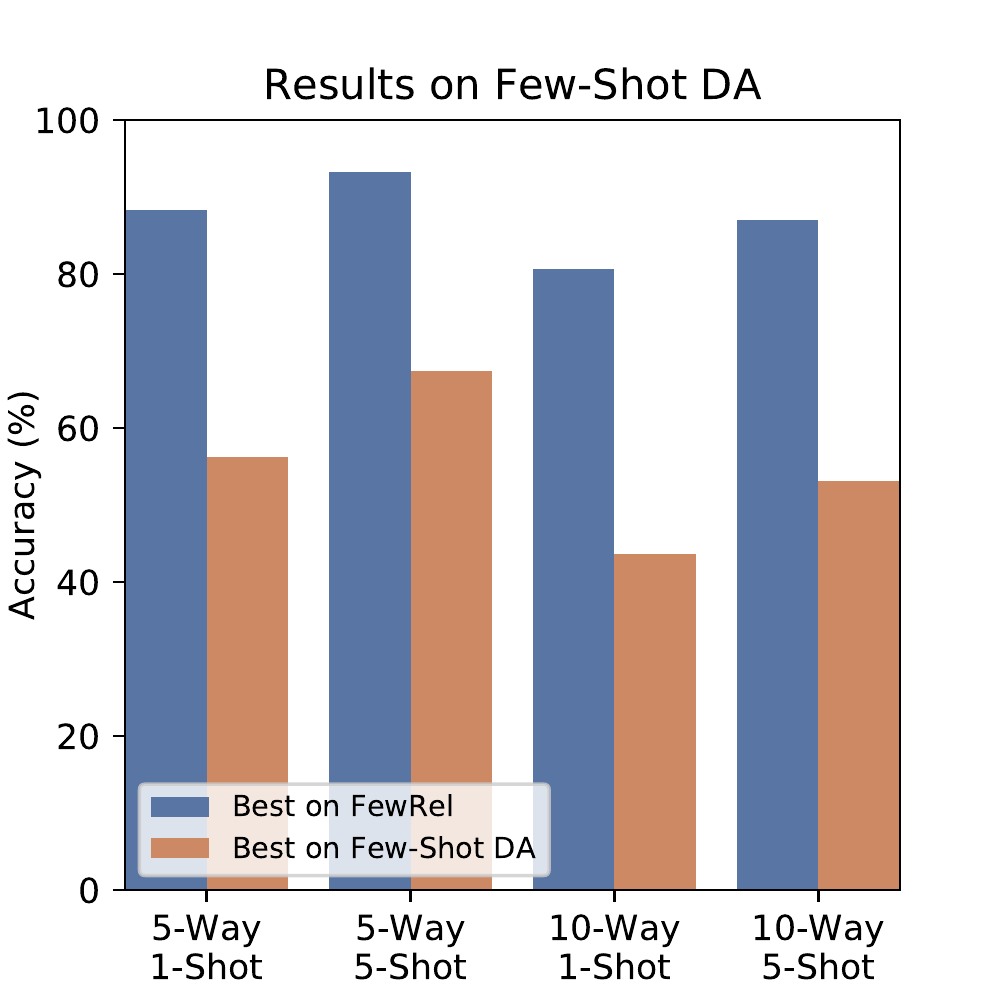}
    \includegraphics[width=0.235\textwidth]{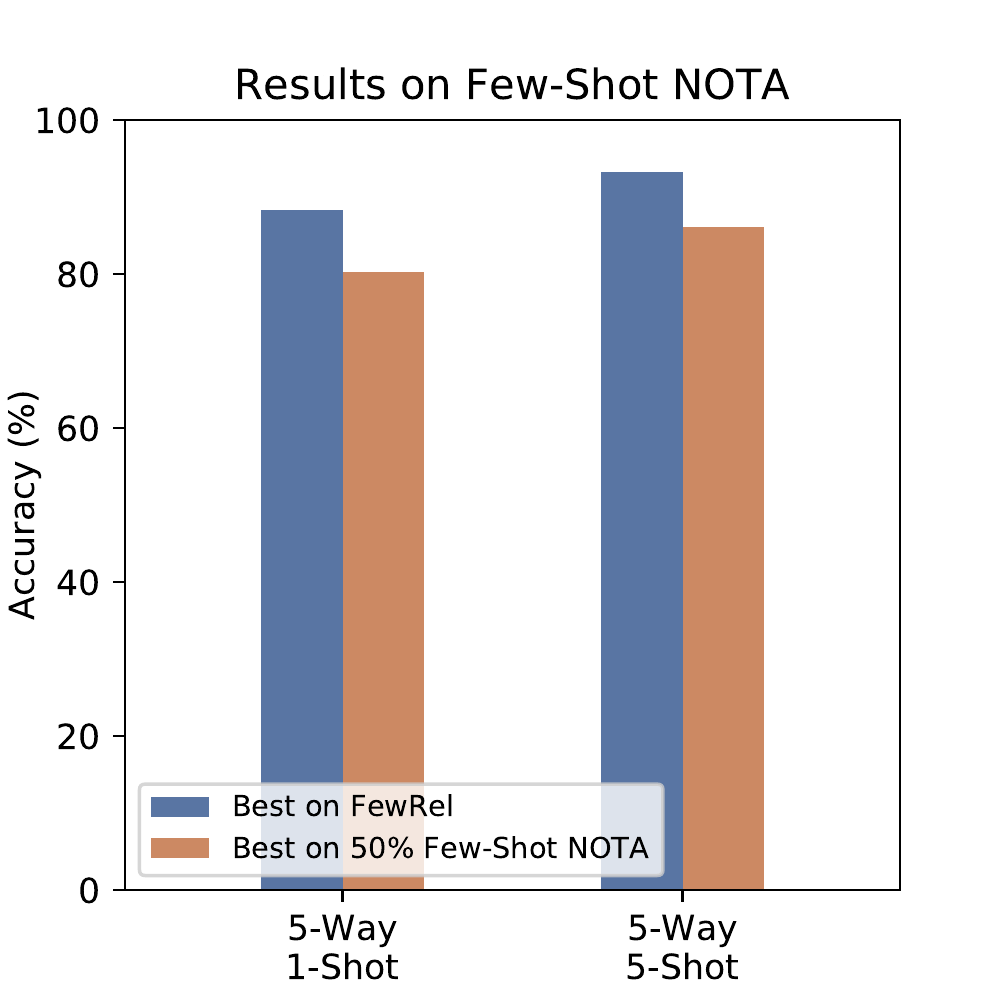}
    \caption{The comparison between the best results of the current models on FewRel, few-shot DA and few-shot NOTA. From the figures we can see that even the state-of-the-art models struggle on the new tasks.}
    % \vspace{-1.0em}
    \label{fig:first}
\end{figure}

\textbf{Few-shot domain adaptation} (few-shot DA) aims to evaluate the abilities of few-shot models to transfer across domains, which is crucial for real-world applications, since the test domains usually lack of annotations and could differ vastly from the training domains. To this end, we construct a new test set sharing great disparities with the original FewRel dataset, and carry out extensive experiments on the state-of-the-art few-shot models and commonly-used domain adaptation methods. Some prior experimental results in Figure~\ref{fig:first} show that even the performance of the most effective methods on FewRel drops drastically on the new test set, proving that few-shot DA is challenging and requires further investigations. 

% \Hao{You should explain this figure in detail. What is the phonomena shown in this figure? What message this figure aims to convey? Figures should be self-contained.}

\textbf{Few-shot none-of-the-above detection} (few-shot NOTA) is an advanced version of the existing $N$-way $K$-shot setting in few-shot learning. The original $N$-way $K$-shot setting samples $N$ classes, as well as $K$ supporting instances and several queries from each class for each test batch, assuming that all queries belong to the sampled $N$ classes. However, in few-shot NOTA, queries could also be none-of-the-above (NOTA), which brings one more option in classification and challenges existing few-shot methods. Considering few-shot NOTA has not yet been widely explored, we propose several solutions based on the state-of-the-art few-shot models and evaluate them with few-shot NOTA setting. Figure~\ref{fig:first} shows that though achieving promising results, there is still a room of improvements for few-shot NOTA.

In the following sections, we first describe the two newly-added challenges in FewRel 2.0, then introduce possible directions for addressing these two issues, and finally present results and observations from our experiments.

\section{FewRel 2.0}
\label{sec:2.0}

\begin{table}[t]
\centering
\small
\scalebox{0.8}{
\begin{tabular}{l|l|p{0.5\columnwidth}}
\toprule
\multicolumn{3}{c}{\textbf{Training Phase (Famous Person from Wikipedia)}}\\
\midrule
\multirow{3}{*}{Supp. Set} & \multirow{1}{*}{(A) date\_of\_birth}     & \emph{\textcolor{blue}{Mark Twain}} was born in \emph{\textcolor{red}{1835}}. \\
\cmidrule(lr){2-3}
& \multirow{2}{*}{(B) place\_of\_birth}    & \emph{\textcolor{blue}{Elvis Presley}} was born in \emph{\textcolor{red}{Memphis, Tennessee}}. \\
\midrule
\multirow{2}{*}{Query}      &   \multirow{2}{*}{(A) or (B) or \textbf{\underline{NOTA}}}      &  \emph{\textcolor{blue}{William Shakespeare}} passed away at age 52 (around \emph{\textcolor{red}{1616}}).\\
\midrule
\midrule
\multicolumn{3}{c}{\textbf{Test Phase (Biomedicine)}}\\
\midrule
\multirow{6}{*}{Supp. Set} & \multirow{2}{*}{(A) may\_treat}   & \emph{\textcolor{blue}{Ribavirin}} remains essential to \emph{\textcolor{red}{Chronic Hepatitis C}} treatment. \\
\cmidrule(lr){2-3}
& \multirow{3}{*}{(B) manifestation\_of}    & Boys with \emph{\textcolor{red}{Prader-Willi syndrome}} often have \emph{\textcolor{blue}{undescended testicle}}. \\
\midrule
\multirow{3}{*}{Query}      &   \multirow{3}{*}{\textbf{\underline{(A)}} or (B) or NOTA}      &  \emph{\textcolor{blue}{Thiabendazole}} was effective in eradicating the \emph{\textcolor{red}{strongyloides infection}}.\\
\bottomrule
\end{tabular}
}
\caption{An example for a 2-way 1-shot scenario, including both few-shot DA and few-shot NOTA. Different colors indicate different entities, \textcolor{blue}{blue} for head entities, and \textcolor{red}{red} for tail entities. For few-shot DA, instances in the training phase and test phase come from different domains. For few-shot NOTA, it requires models to detect the none-of-the-above (NOTA) relation.}
\label{tab:example}
\end{table}

\subsubsection*{Formulation for $N$-Way $K$-Shot Setting}

%\label{section:nk}

The original FewRel task adopts the $N$-way $K$-shot setting. The whole dataset is divided into training, validation and test subsets, which have no intersection in relation types. Models are evaluated with batches sampled from the test set, each of which consists of $(\mathcal{R}, \mathcal{S}, x, r)$, where $\mathcal{R}=\{r_1,r_2,...,r_N\}$ is the sampled relation set, $r\in \mathcal{R}$ is the correct relation label for the query $x$, and $\mathcal{S}$ is the supporting set containing $K$ instances for each relation,
\begin{equation}
    % \small
    \mathcal{S}=\{(x_{r_i}^j, r_i)\},1\leq i\leq N,1\leq j\leq K.
\end{equation}
Models should predict the relation label $y\in \mathcal{R}$ for the query instance $x$ based on the given $\mathcal{S}$ and $\mathcal{R}$. Both of the following two challenges are based on this $N$-way $K$-shot setting.

\subsubsection*{Few-Shot Domain Adaptation}

%\label{section:da}

Both the training and test sets of the original FewRel dataset are constructed by manually annotating the distantly supervised~\cite{bunescu2007learning, mintz2009distant} results on Wikipedia corpus and Wikidata~\cite{vrandevcic2014wikidata} knowledge bases. In other words, they are from the same domain, yet in a real-world scenario, we might train models on one domain and perform few-shot learning on a different one. For example, we may train models on Wikipedia, which has large amounts of data and adequate annotations, and then perform few-shot learning on some domains suffering data sparsity, like literature, finance and medicine. Note that, not only do these corpora differ vastly from each other in morphology and syntax, but there are wide disparities between the relation sets defined on these domains as well, which makes transferring knowledge across different domains more challenging.

To explore few-shot DA, we construct a new test set by aligning PubMed~\footnote{\url{https://www.ncbi.nlm.nih.gov/pubmed/}}, a database containing large amounts of biomedical literature, with UMLS~\footnote{UMLS represents the Unified Medical Language System$^\circledR$, which is the trademark of U.S. National Library of Medicine.}, a large-scale knowledge base in the biomedical sciences. Then we let the annotators classify whether each instance we get from the distant supervision is correct. Every sentence is assigned to at least two annotators, and if their annotation results do not agree with each other, the third annotator is assigned. In the end, we gather a valid dataset with 25 relations and 100 instances for each relation.

For few-shot DA, we adopt the original FewRel training set for training, and the newly-annotated dataset for test, as shown in Table \ref{tab:example}. Besides, we use SemEval-2010 task 8 dataset~\cite{hendrickx2009semeval} as the validation set, since both the corpora and the schema of SemEval-2010 task 8 are in different domains from the original FewRel dataset and the newly-annotated test set.

\subsubsection*{Few-Shot None-of-the-Above Detection}

\label{section:nota}

In a $N$-way $K$-shot, all queries are assumed to be in the given relation set, yet sentences expressing no specific relations or relations not in the given set should also be taken into consideration, for they make up the vast majority of text. This calls for the none-of-the-above (NOTA) relation, which indicates that the query instance does not express any of the given relations. Though it is common in some conventional classification tasks, where NOTA is usually regarded as an extra class, detecting NOTA could be hard in few-shot learning, because the given relation sets are not fixed so that the NOTA relation requires to cover a different semantic space each time. An example of NOTA is given in Table \ref{tab:example}.

We formalize few-shot NOTA based on the $N$-way $K$-shot setting. For the query instance $x$, the correct relation label becomes $r\in \{r_1,r_2,...,r_N,\texttt{NOTA} \}$ rather than $r \in \{r_1,r_2,...,r_N\}$. We use the parameter NOTA rate to describe the proportion of NOTA queries during the whole test phase. For example, $0\%$ NOTA rate means no queries are NOTA and $50\%$ NOTA rate means half of the queries have the label \texttt{NOTA}. 

The NOTA queries are sampled from those relations outside the given $N$ relations. To be more specific, denoting the whole test set as $\mathcal{D}_{\texttt{test}}$, the set containing all instances in the relation set $\mathcal{R}$ as $\mathcal{D}_{\mathcal{R}}$ and the NOTA rate as $\alpha$, $\alpha$ of the query instances (NOTA queries) are from $\mathcal{D_{\texttt{test}}}\setminus \mathcal{D}_{\mathcal{R}}$ and $1-\alpha$ of the instances are from $\mathcal{D}_{\mathcal{R}}$.
%In few-shot NOTA, the query instance has the probability of NOTA rate that comes from $\mathcal{D_{\texttt{test}}}\setminus \mathcal{D}_{\mathcal{R}}$ (NOTA queries), otherwise it comes from $\mathcal{D}_{\mathcal{R}}$.

Note that during the test phase, all the queries are from the test set, though models can sample instances from the training set as supporting instances for NOTA relation (this method is described explicitly in Section \ref{section:methodnota}). Also note that to better demonstrate the effects of the NOTA relation, we use the original FewRel dataset for few-shot NOTA, instead of the new test set, which can get rid of the influence of domain adaptation.

%We do not limit how models utilize data from the training set. 

%We do not limit the NOTA rate during the training phase.
%To simulate different few-shot scenes with none-of-the-above (NOTA) queries, we set 4 different levels of NOTA rate: $0\%$, $15\%$, $30\%$, $50\%$, where $0\%$ represents there are no NOTA queries and $50\%$ means half of the queries are NOTA.

\section{Approaches for Few-Shot DA}

\label{sec:methodda}

Many efforts have been devoted for domain adaptation, like subspace mapping~\cite{pan2010domain, fernando2013unsupervised}, finding domain-invariant spaces~\cite{baktashmotlagh2013unsupervised, ganin2016domain}, feature augmentation~\cite{blitzer2006domain} and minimax estimators~\cite{provost2001robust}. Among them, adversarial training~\cite{goodfellow2014explaining,ganin2016domain,wang2018adversarial} has been proved to be efficient in finding domain-invariant features.
It is a game process between an encoder and a discriminator, where the encoder tries to generate domain-invariant features while the discriminator tries to tell which domain the features are from.

%We incorporate with adversarial training in experiments and achieve promising results, yet there is still huge space for improvement. 
Here we follow the adversarial training setting in~\newcite{wang2018adversarial}, where a two-layer perceptron network is used as the discriminator. While training the few-shot learning task, we feed the sentence encoder $\bm{E}$ and the discriminator $\bm{D}$ with the corpora from the training domain and the test domain, and optimize the min-max game,
\begin{equation}
% \small
\begin{aligned}
    \min_{\theta_{\bm{E}}} \max_{\theta_{\bm{D}}} \sum_{x\in\mathcal{C}_0} \log [\bm{D}(\bm{E}(x))]_0 \\ +\sum_{x\in\mathcal{C}_1} \log [\bm{D}(\bm{E}(x))]_1,
\end{aligned}
\end{equation}
where $[\cdot]_i$ is the $i$-th element of the vector, $\mathcal{C}_0$ is the training corpus and $\mathcal{C}_1$ is the test corpus.

\section{Approaches for Few-Shot NOTA}

\label{section:methodnota}

%To the best of our knowledge, there are no research in none-of-the-above detection.
A simple way to handle NOTA is to regard it as an extra class in the $N$-way $K$-shot setting.
To be more specific, we can sample instances outside the $N$ relations as the supporting data of NOTA, and perform the $(N+1)$-way $K$-shot learning.
As compared to the current methods ignoring NOTA, this approach does not bring much improvements, since the supporting data for NOTA actually belong to several different relations and are scattered in the feature space, making it hard to perform classification.

\begin{table*}[t]
    \centering
    \small
    \renewcommand\arraystretch{1.2}
    \scalebox{1.0}{
    \begin{tabular}{l|c|c|c|c}
        \toprule
         \multicolumn{1}{c|}{\multirow{2}{*}{\bf Model}} & \multicolumn{2}{c|}{\bf 5-Way 1-Shot} & \multicolumn{2}{c}{\bf 5-Way 5-Shot}\\
         \cmidrule{2-5}
         &\bf On 1.0 &\bf On 2.0 & \bf On 1.0 & \bf On 2.0 \\
         \midrule
         %Meta Network (CNN)          & $64.46\pm0.54$ & $00.00\pm0.00$ & $80.57\pm0.48$ & $00.00\pm0.00$\\
         GNN (CNN)                   & $66.23\pm0.75$ & $27.94\pm0.03$ & $81.28\pm0.62$ & $29.33\pm0.11$ \\
         %Prototypical Network (CNN)  & $69.20\pm0.20$ & $35.09\pm0.10$ & $84.79\pm0.16$ & $49.37\pm0.10$\\
         Proto (CNN)  & $74.52\pm0.07$ & $35.09\pm0.10$ & $88.40\pm0.06$ & $49.37\pm0.10$ \\
         Proto-ADV (CNN) & $70.28\pm0.15$ & $42.21\pm0.09$ & $84.63\pm0.07$& $58.71\pm0.06$ \\
         Proto (BERT) & $80.68\pm0.28$ & $40.12\pm0.19$ & $89.60\pm0.09$ & $51.50\pm0.29$\\
         Proto-ADV (BERT) & $73.35\pm0.95$ &$41.90\pm0.44$ &$82.30\pm0.53$ &$54.74\pm0.22$ \\
         BERT-PAIR                    & $88.32\pm0.64$ & $56.25\pm0.40$ & $93.22\pm0.13$ & $67.44\pm0.54$\\
         \midrule
         \midrule
         \multicolumn{1}{c|}{\multirow{2}{*}{\bf Model}} & \multicolumn{2}{c|}{\bf 10-Way 1-Shot} & \multicolumn{2}{c}{\bf 10-Way 5-Shot}\\
         \cmidrule{2-5}
         &\bf On 1.0 &\bf On 2.0 & \bf On 1.0 & \bf On 2.0 \\
         \midrule
         %Meta Network (CNN)          & $64.46\pm0.54$ & $00.00\pm0.00$ & $80.57\pm0.48$ & $00.00\pm0.00$\\
         GNN (CNN)                   & $46.27\pm0.80$ & $16.44\pm0.04$ & $64.02\pm0.77$ & $18.26\pm0.03$\\
         %Prototypical Network (CNN)  & $69.20\pm0.20$ & $35.09\pm0.10$ & $84.79\pm0.16$ & $49.37\pm0.10$\\
         Proto (CNN)  & $62.38\pm0.06$ & $22.98\pm0.05$ & $80.45\pm0.08$ & $35.22\pm0.06$\\
         Proto-ADV (CNN) & $56.34\pm0.08$& $28.91\pm0.10$ & $74.67\pm0.12$ & $44.35\pm0.09$ \\
         Proto (BERT) & $71.48\pm0.15$ & $26.45\pm0.10$ & $82.89\pm0.11$ & $36.93\pm0.01$\\
         Proto-ADV (BERT) & $61.49\pm0.69$ & $27.36\pm0.50$ & $72.60\pm0.38$ & $37.40\pm0.36$\\
         BERT-PAIR                    &                    $80.63\pm0.17$ & $43.64\pm0.46$ & $87.02\pm 0.12$ & $53.17\pm0.09$ \\
        \bottomrule
    \end{tabular}}
    \caption{Accuracies (\%) on few-shot DA. ``On 1.0'' represents the results on the original FewRel dataset and ``On 2.0'' represents the results on the new test set. The models with ``-ADV'' use adversarial training described in Section \ref{sec:methodda}.}% All the few-shot models struggle on the new test set, proving few-shot DA is a challenging task for existing methods.}% ``CNN'' and ``BERT'' are encoders used in the models.}
    \label{table:da}
    \vspace{-0.6em}
\end{table*}

To better address few-shot NOTA, we propose a model named \textbf{BERT-PAIR} based on the sequence classification model in BERT~\cite{devlin2018bert}. We pair each query instance with all the supporting instances, concatenate each pair as one sequence, and send the concatenated sequence to the BERT sequence classification model to get the score of the two instances expressing the same relation. Denote the BERT model as $\bm{B}$, the query instance as $x$ and the paired supporting instance as $x_{r}^j$ (the $j$-th supporting instance for the relation $r$), 
$\bm{B}(x, x_{r}^j)$ outputs a two-element vector corresponding to scores of the pair sharing the same relation and not sharing the same relation. 
The probability over each relation in the few-shot scenario, including NOTA, is addressed as follows,
\begin{equation}
    % \small
    p(y=r|x)=\frac{\exp(o_r)}{\sum_{r'\in \mathcal{R}}\exp(o_{r'})}, r\in \mathcal{R}
\end{equation}
where $y$ is the predicted label and $\mathcal{R}=\{r_1,...,r_N,\texttt{NOTA}\}$ is the relation set including NOTA. For $r\in \{r_1,...,r_N\}$, $o_r$ is calculated by averaging,
\begin{equation}
% \small
o_r=\frac{1}{K}\sum_{j=1}^{K} [\bm{B}(x, x_{r}^j)]_1.
\end{equation}
The score for NOTA $o_{\texttt{NOTA}}$ is calculated by the equation,
\begin{equation}
% \small
o_{\texttt{NOTA}}=\min_{r\in \{r_1,...,r_N\}}{\frac{1}{K}\sum_{j=1}^{K} [\bm{B}(x, x_{r}^j)]_0}.
\end{equation}
Then we can treat NOTA the same as other relations and optimize the model with the cross entropy loss, which is commonly-used in few-shot learning and other classification tasks. 

\section{Experiments}
\label{section:exp}

\subsection{Baseline Models for Few-Shot Learning}

%We select \textbf{GNN}~\cite{garcia2017few} and Prototypical Networks (\textbf{Proto})~\cite{snell2017prototypical}, the two best methods in~\cite{han2018fewrel}, as our baseline models. Besides the CNN encoder used in~\cite{han2018fewrel}, we also adopt Bidirectional Encoder Representations from Transformers ({BERT}) since it achieves state-of-the-arts in multiple tasks~\cite{devlin2018bert}. For all models and encoders, we follow the parameter settings from~\cite{han2018fewrel} and~\cite{devlin2018bert}. %Compared to other encoders like RNN, LSTM and GRU, CNN is far more efficient and BERT achieves significantly better results. %Besides, we evaluate the adversarial training technique on the domain adaptation test set and the proposed \textbf{BERT-PAIR} model on both challenges. 

We pick the two best models from the results in \newcite{han2018fewrel}, \textbf{GNN}~\cite{satorras2018few} and \textbf{Prototypical Networks}~\cite{snell2017prototypical}, as our baseline models. As for the encoders, besides the CNN encoder used in \newcite{han2018fewrel}, we also adopt BERT since it achieves the state-of-the-arts in multiple tasks~\cite{devlin2018bert}. For all models and encoders, we follow the parameter settings from~\newcite{han2018fewrel} and~\newcite{devlin2018bert}. 

\begin{comment}
We pick two best models from results in \cite{han2018fewrel}, GNN and Prototypical Networks, as our baseline models.

\paragraph{GNN} GNN~\cite{satorras2018few} tackles few-shot learning with graph neural networks, taking the embeddings of supporting and query instances as nodes in the graph. Information of instances propagates in the graph and finally we expect the query nodes gather the information they need for classification.

\paragraph{Prototypical Networks} Prototypical Networks (Proto)~\cite{snell2017prototypical} calculate the prototype for each class by averaging all the representations of instances inside the class, and classify each query by computing the Euclidean distance between it and the prototype for each class. 

\paragraph{} Besides the CNN encoder used in \cite{han2018fewrel}, we also adopt Bidirectional Encoder Representations from Transformers ({BERT}) since it achieves state-of-the-arts in multiple tasks~\cite{devlin2018bert}. For all models and encoders, we follow the parameter settings from~\cite{han2018fewrel} and~\cite{devlin2018bert}. 
\end{comment}

\begin{figure}[t]
    \centering
    \includegraphics[width=0.235\textwidth]{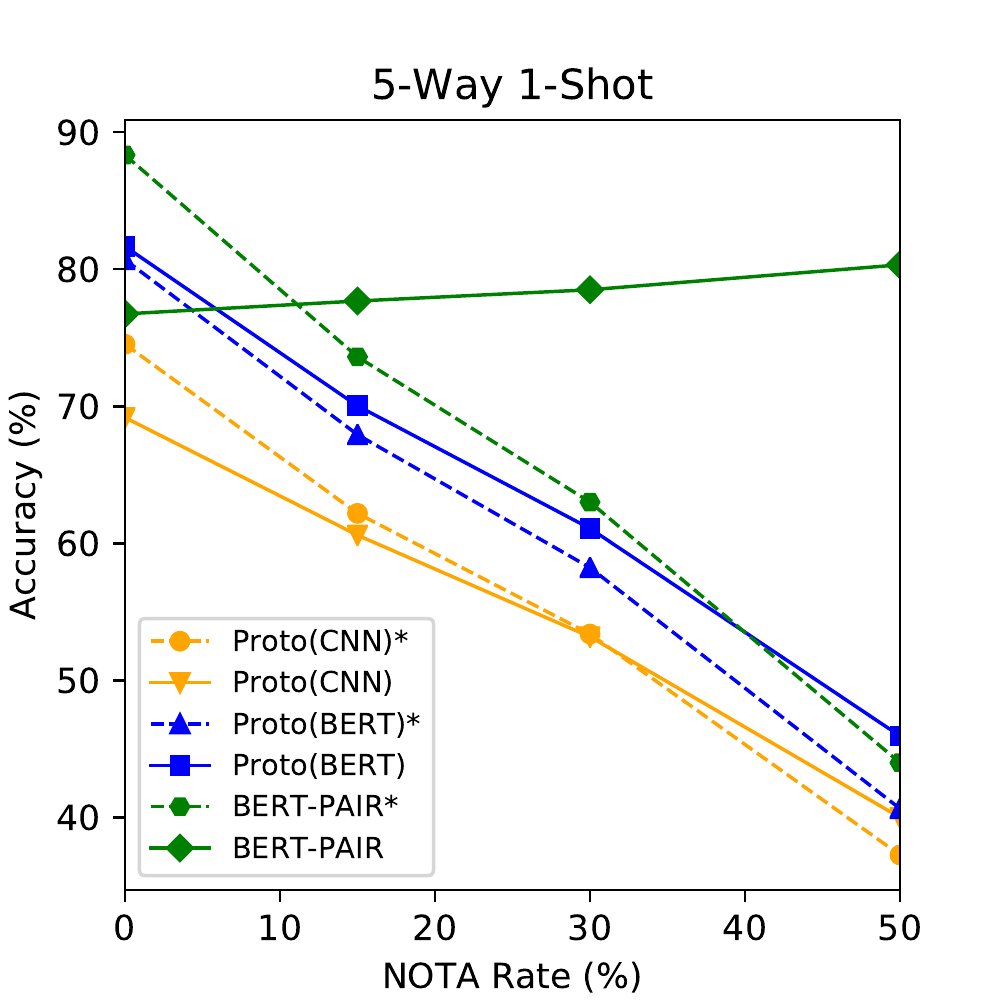}
    \includegraphics[width=0.235\textwidth]{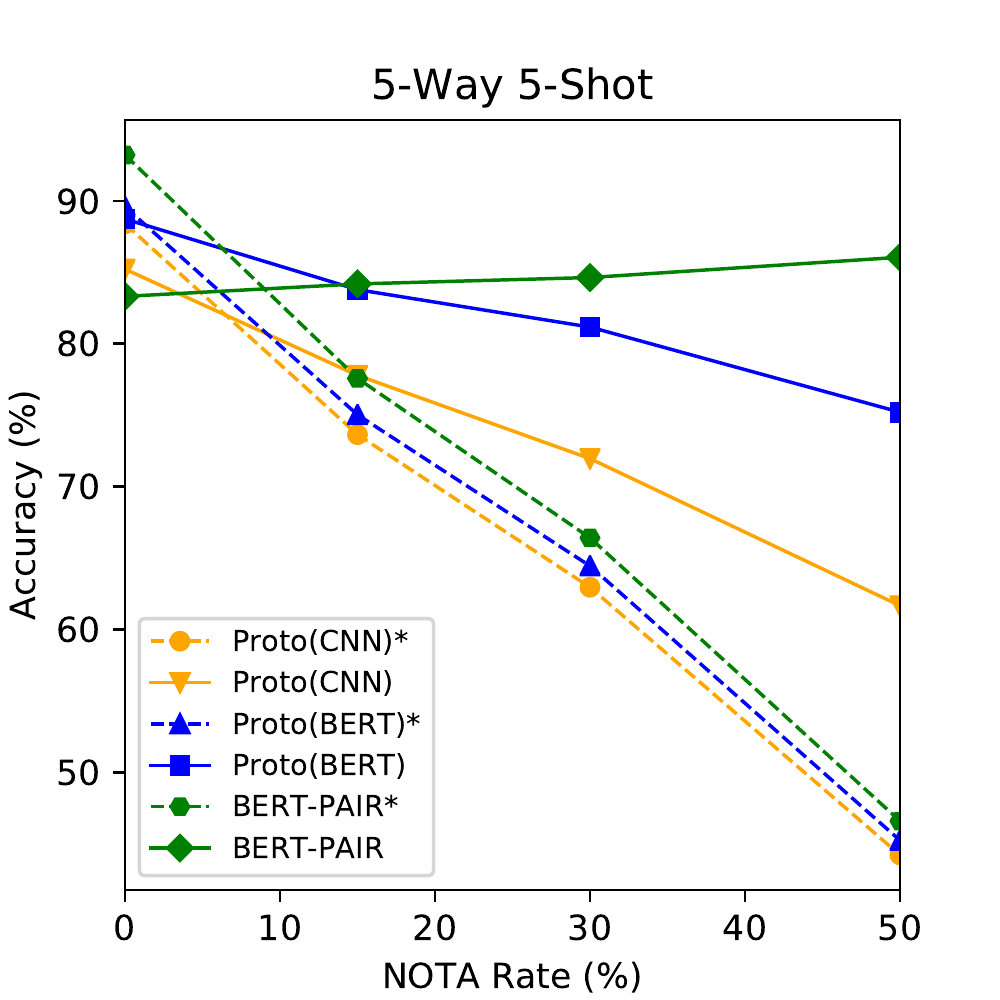}
    \caption{5-way $K$-shot results under different NOTA rates. Models with * simply ignore the NOTA setting and assume all queries can be classified as one of the $N$ relations.}%Even for the best model (BERT-PAIR) under few-shot NOTA, there is still a performance gap to the state-of-the-art of 0\% NOTA setting.}%As the figure suggests, though not the best one when there are no NOTA queries, BERT-SEQ achieves stable and state-of-the-art performance when NOTA rate rises.}
    \label{fig:nota}
    \vspace{-0.1em}
\end{figure}

\begin{table*}[t]
    \centering
    \small
    \renewcommand\arraystretch{1.2}
    \scalebox{1.0}{
    \begin{tabular}{l|c|c|c|c}
    \toprule
    \multicolumn{1}{c|}{\multirow{2}{*}{\bf Model}} & \multicolumn{4}{c}{\bf 5-Way-1-Shot} \\
    \cmidrule{2-5}
    %&  0\% NOTA  & 15\% NOTA & 30\% NOTA & 50\% NOTA &  0\% NOTA  & 15\% NOTA & 30\% NOTA & 50\% NOTA \\
    &  {\bf 0\% NOTA}  & {\bf 15\% NOTA} & {\bf 30\% NOTA} & {\bf 50\% NOTA} \\
    \midrule
    %Proto (CNN)  & $$ & $$$ & $$ & $$\\
    Proto (CNN)*   & $74.52\pm0.07$ & $62.18\pm0.22$ &$53.38\pm0.14$  & $37.26\pm0.04$ \\
    Proto (CNN)    & $69.17\pm0.07$ & $60.59\pm0.05$ & $53.18\pm0.12$ & $40.00\pm0.10$\\
    Proto (BERT)*  & $80.68\pm0.28$ & $67.92\pm0.31$ & $58.22\pm0.20$ & $40.64\pm0.14$ \\
    Proto (BERT)   & $81.65\pm0.97$ & $70.02\pm0.23$ & $61.08\pm0.28$ & $45.94\pm0.50$ \\
    BERT-PAIR*      & $88.32\pm0.64$ & $73.60\pm0.51$ & $63.00\pm0.47$ & $43.99\pm0.09$ \\
    BERT-PAIR       & $76.73\pm0.55$ & $77.67\pm0.14$ & $78.49\pm0.21$ & $80.31\pm0.12$ \\
    \end{tabular}}
    \scalebox{1.0}{
    \begin{tabular}{l|c|c|c|c}
    \midrule
    \midrule
    \multicolumn{1}{c|}{\multirow{2}{*}{\bf Model}} & \multicolumn{4}{c}{\bf 5-Way-5-Shot} \\
    \cmidrule{2-5}
    %&  0\% NOTA  & 15\% NOTA & 30\% NOTA & 50\% NOTA &  0\% NOTA  & 15\% NOTA & 30\% NOTA & 50\% NOTA \\
    &  {\bf 0\% NOTA}  & {\bf 15\% NOTA} & {\bf 30\% NOTA} & {\bf 50\% NOTA} \\
    \midrule
    %Proto (CNN)  & $$ & $$$ & $$ & $$\\
    Proto (CNN)*  & $88.40\pm0.06$ & $73.64\pm0.11$&$62.95\pm0.12$ & $44.20\pm0.05$\\
    Proto (CNN)    &$85.23\pm0.07$ & $77.79\pm0.03$ & $71.96\pm0.14$ & $61.66\pm0.08$\\
    Proto (BERT)*  &  $89.60\pm0.09$ & $75.03\pm0.17$ & $64.44\pm0.18$ & $45.22\pm0.03$ \\
    Proto (BERT)   &  $88.74\pm0.83$ & $83.79\pm0.44$ & $81.17\pm0.48$ & $75.21\pm0.52$\\
    BERT-PAIR*      &  $93.22\pm0.13$ & $77.58\pm0.42$ & $66.41\pm0.24$ &$46.58\pm0.09$ \\
    BERT-PAIR       &  $83.32\pm0.38$ & $84.19\pm0.46$ & $84.64\pm0.13$ & $86.06\pm0.43$\\
    \bottomrule
    \end{tabular}}
    \caption{Accuracies (\%) on few-shot NOTA. Models with * simply ignore the NOTA setting and assume all queries can be classified as one of the given relations.}
    \label{table:nota}
    \vspace{-1.0em}
\end{table*}

\subsection{Evaluation Results on Few-Shot DA}

Table~\ref{table:da} demonstrates the evaluation results of few-shot DA on the existing FewRel test set and the new test set. Besides the baselines, we also evaluate Prototypical Networks with adversarial training described in Section~\ref{sec:methodda} and our proposed BERT-PAIR model in Section~\ref{section:methodnota}. We get three observations from the results: 

(1) All few-shot models suffer dramatic performance falls when tested on a different domain.

(2) Adversarial training does improve the results on the new test domain, yet still has large space for growth.

(3) BERT-PAIR outperforms all other few-shot models on both 1.0 and 2.0 test set. 

Besides, to see where the growth boundary is, we split $10$ relations, $1,000$ instances out of the 2.0 test set and add them to the training set, then train and evaluate BERT-PAIR on the new data. We get $72.30\%$ for 5-way 1-shot and $80.50\%$ for 5-way 5-shot, $16$ and $13$ points higher than the current best results. Note that only $1,000$ training instances can lead to such an enormous gap, indicating that there is still a huge room for improvements.

%, proving that adapting to a new domain is challenging. 
%We can also see that adversarial training does improve the results on the new test set. Besides, BERT-PAIR outperforms all other models on 1.0 and 2.0 test set. Though adversarial training and BERT-PAIR model achieve promising results in few-shot domain adaptation, there are still huge gaps between results on the original domain and on the new domain, which calls for further research in this topic. 

%Note that the proposed BERT-PAIR model not only achieves the best result on both the 1.0 and 2.0 test sets, but it also suffers less than other models on the new domain.% (BERT-PAIR only drops by 39\% of the original results on 10-Way 5-Shot setting while Proto (BERT) drops by 55\%). 

%To evaluate the power of adversarial training, we perform the technique on the Prototypical Networks and compare the results with and without the adversarial training. The reason we do not train BERT-PAIR with adversarial approach is that BERT-PAIR directly outputs the features for the concatenation of the two sentences, instead of the feature for each sentence. From Table \ref{table:adv-pubmed} we can see that adversarial training does improve the results of Prototypical Networks with both encoders on the new test set.

\subsection{Evaluation Results on Few-Shot NOTA}

We evaluate Prototypical Networks with the naive NOTA solution described in Section~\ref{section:methodnota} and BERT-PAIR under the NOTA setting. All models are trained given $50\%$ NOTA queries and tested under four different NOTA rates: $0\%$, $15\%$, $30\%$, $50\%$. To show how accuracy falls if ignoring the NOTA relation, we also demonstrate the results of models without considering NOTA (marked with * in Figure~\ref{fig:nota}). We demonstrate the evaluation results in Figure~\ref{fig:nota}. For detailed numbers of results on few-shot NOTA, please refer to Table~\ref{table:nota}. From Figure~\ref{fig:nota} we can conclude that: 

(1) Treating NOTA as the $N+1$ relation is beneficial for handling Few-Shot NOTA, though the results still fall fast when the NOTA rate increases. 

(2) BERT-PAIR works better under the NOTA setting for its binary-classification style model, 
and stays stable %, even increases a little 
with rising NOTA rate. 
%(3) Compared to ignoring NOTA, both Proto and BERT-PAIR considering NOTA decrease under the conventional setting (0\% NOTA), which is understandable since taking NOTA means having an additional option for classification. 

(3) Though BERT-PAIR achieves promising results, huge gaps still exist between the conventional ($0\%$ NOTA rate) and NOTA settings (gaps of $8$ points for 5-way 1-shot and $7$ points for 5-way 5-shot with $50\%$ NOTA rate), which calls for further research to address the challenge.

\section{Conclusion}

In this paper, we propose FewRel 2.0, a more challenging few-shot relation classification task with a new test set from the biomedical domain and the none-of-the-above setting. The purpose of the new task is to explore two aspects which are ignored in the previous work: few-shot domain adaptation (few-shot DA) and few-shot none-of-the-above detection (few-shot NOTA). Extensive experiments demonstrate that the existing state-of-the-art few-shot models struggle on the new task. We also point out some possible directions to handle these two issues, implement several new models and evaluate them with the new task. Though achieving promising improvements, these commonly-used techniques are still not the satisfactory solutions for few-shot DA and few-shot NOTA, which requires further explorations in these two real-world challenges. 

\section*{Acknowledgments}

This work is supported by the National Natural Science Foundation of China (NSFC No. 61572273, 61661146007) and Tsinghua University Initiative Scientific Research Program (20151080406). This work is also supported by the Pattern Recognition Center, WeChat AI, Tencent Inc. Han and Gao are supported by 2018 and 2019 Tencent Rhino-Bird Elite Training Program respectively. Gao is also supported by Tsinghua University Initiative Scientific Research Program. We also thank Xiaozhi Wang for his insightful ideas and suggestions.

\bibliography{emnlp-ijcnlp-2019}
\bibliographystyle{acl_natbib}

\end{document}

\maketitle

\appendix
\section{Evaluations on Few-Shot NOTA}

This is the detailed evaluation results on few-shot none-of-the-above detection. 

\begin{table}[h]
    \centering
    \small
    \renewcommand\arraystretch{1.2}
    \scalebox{0.7}{
    \begin{tabular}{l|c|c|c|c}
    \toprule
    \multicolumn{1}{c|}{\multirow{2}{*}{\bf Model}} & \multicolumn{4}{c}{\bf 5-Way-1-Shot} \\
    \cmidrule{2-5}
    %&  0\% NOTA  & 15\% NOTA & 30\% NOTA & 50\% NOTA &  0\% NOTA  & 15\% NOTA & 30\% NOTA & 50\% NOTA \\
    &  {\bf 0\% NOTA}  & {\bf 15\% NOTA} & {\bf 30\% NOTA} & {\bf 50\% NOTA} \\
    \midrule
    %Proto (CNN)  & $$ & $$$ & $$ & $$\\
    Proto (CNN)*   & $74.52\pm0.07$ & $62.18\pm0.22$ &$53.38\pm0.14$  & $37.26\pm0.04$ \\
    Proto (CNN)    & $69.17\pm0.07$ & $60.59\pm0.05$ & $53.18\pm0.12$ & $40.00\pm0.10$\\
    Proto (BERT)*  & $80.68\pm0.28$ & $67.92\pm0.31$ & $58.22\pm0.20$ & $40.64\pm0.14$ \\
    Proto (BERT)   & $81.65\pm0.97$ & $70.02\pm0.23$ & $61.08\pm0.28$ & $45.94\pm0.50$ \\
    BERT-PAIR*      & $88.32\pm0.64$ & $73.60\pm0.51$ & $63.00\pm0.47$ & $43.99\pm0.09$ \\
    BERT-PAIR       & $76.73\pm0.55$ & $77.67\pm0.14$ & $78.49\pm0.21$ & $80.31\pm0.12$ \\
     \bottomrule
    \end{tabular}}
    \scalebox{0.7}{
    \begin{tabular}{l|c|c|c|c}
    \toprule
    \multicolumn{1}{c|}{\multirow{2}{*}{\bf Model}} & \multicolumn{4}{c}{\bf 5-Way-5-Shot} \\
    \cmidrule{2-5}
    %&  0\% NOTA  & 15\% NOTA & 30\% NOTA & 50\% NOTA &  0\% NOTA  & 15\% NOTA & 30\% NOTA & 50\% NOTA \\
    &  {\bf 0\% NOTA}  & {\bf 15\% NOTA} & {\bf 30\% NOTA} & {\bf 50\% NOTA} \\
    \midrule
    %Proto (CNN)  & $$ & $$$ & $$ & $$\\
    Proto (CNN)*  & $88.40\pm0.06$ & $73.64\pm0.11$&$62.95\pm0.12$ & $44.20\pm0.05$\\
    Proto (CNN)    &$85.23\pm0.07$ & $77.79\pm0.03$ & $71.96\pm0.14$ & $61.66\pm0.08$\\
    Proto (BERT)*  &  $89.60\pm0.09$ & $75.03\pm0.17$ & $64.44\pm0.18$ & $45.22\pm0.03$ \\
    Proto (BERT)   &  $88.74\pm0.83$ & $83.79\pm0.44$ & $81.17\pm0.48$ & $75.21\pm0.52$\\
    BERT-PAIR*      &  $93.22\pm0.13$ & $77.58\pm0.42$ & $66.41\pm0.24$ &$46.58\pm0.09$ \\
    BERT-PAIR       &  $83.32\pm0.38$ & $84.19\pm0.46$ & $84.64\pm0.13$ & $86.06\pm0.43$\\
    \bottomrule
    \end{tabular}}
    \caption{Accuracies (\%) on few-shot NOTA. Models with * ignore the NOTA setting and regard that all queries belong to one of the given relations.}
    \label{table:nota}
\end{table}

\section{Few-Shot DA Test Set}

\begin{table}[t]
    \centering
    \small
    \renewcommand\arraystretch{1.2}
    %\scalebox{0.7}{
    \begin{tabular}{l}
    \toprule
\textbf{Relation Name}\\
\midrule
is\_associated\_anatomy\_of\_gene\_product\\
finding\_site\_of\\
has\_structural\_class\\
disposition\_of\\
may\_be\_treated\_by\\
chemotherapy\_regimen\_has\_component\\
has\_method\\
procedure\_site\_of\\
gene\_product\_has\_biochemical\_function\\
manifestation\_of\\
process\_involves\_gene\\
part\_of\\
may\_be\_prevented\_by\\
disease\_has\_associated\_anatomic\_site\\
disease\_has\_normal\_cell\_origin\\
biological\_process\_involves\_gene\_product\\
inheritance\_type\_of\\
is\_normal\_tissue\_origin\_of\_disease\\
ingredient\_of\\
is\_primary\_anatomic\_site\_of\_disease\\
gene\_found\_in\_organism\\
occurs\_in\\
causative\_agent\_of\\
classified\_as\\
gene\_plays\_role\_in\_process\\
\bottomrule
    \end{tabular}
    %}
    \caption{25 relations in few-shot DA test set.}
    \label{tab:relation}
\end{table}

We adopt 25 relations defined in UMLS\footnote{UMLS represents the Unified Medical Language System$^\circledR$, which is the trademark of U.S. National Library of Medicine.}, which are listed in Table \ref{tab:relation}. We also provide the data and annotations of our new test set in the supplementary materials.